# NLP Meets RNA:
# Unsupervised Embedding Learning for Ribozymes with Word2Vec

Andrew Kean Gao
Stanford University

## Abstract

Ribozymes, RNA molecules with distinct 3-D structures and catalytic activity, have widespread applications in synthetic biology and therapeutics. However, relatively little research has focused on leveraging deep learning to enhance our understanding of ribozymes. This study implements Word2Vec, an unsupervised learning technique for natural language processing, to learn ribozyme embeddings. Ribo2Vec was trained on over 9,000 diverse ribozymes, learning to map sequences to 128 and 256-dimensional vector spaces. Using Ribo2Vec, sequence embeddings for five classes of ribozymes – hatchet, pistol, hairpin, hovlinc, and twister sister – were calculated. Principal component analysis demonstrated the ability of these embeddings to distinguish between ribozyme classes. Furthermore, a simple SVM classifier trained on ribozyme embeddings showed promising results in accurately classifying ribozyme types. Our results suggest that the embedding vectors contained meaningful information about ribozymes. Interestingly, 256-dimensional embeddings behaved similarly to 128-dimensional embeddings, suggesting that a lower dimension vector space is generally sufficient to capture ribozyme features. This approach demonstrates the potential of Word2Vec for bioinformatics, opening new avenues for ribozyme research. Future research includes using a Transformer-based method to learn RNA embeddings, which can capture long-range interactions between nucleotides.

We created a public web server for Ribo2Vec: https://ribo2vec.sites.stanford.edu/

# Background:

Ribozymes, also known as catalytic RNAs, are RNA molecules that catalyze specific biochemical reactions, similar to protein-based enzymes [1]. Ribozymes are a type of non-coding RNA, a broad class that includes microRNAs, circRNAs, lncRNAs, and more [2]. They play a crucial role in various cellular functions, including RNA splicing, translation, and viral replication. Ribozymes are interesting because like proteins, they form specific 3-D structures that are vital to their catalytic functions. This is not common in RNA (which do have some structure, but not catalytic activity). Ribozymes are classified into several types, including hatchet, pistol, hairpin, hovlinc, and twister sister, each with unique structural and functional characteristics [3]. The study of ribozymes is of importance because of their applications in synthetic biology, therapeutics, and as tools for molecular biology [4-5]. Despite their importance, our understanding of ribozymes is still limited, particularly in terms of their three-dimensional structures and the relationship between their sequence, structure, and function. Relatively little research has been conducted on ribozymes when compared to proteins, which have been a focus of current research [6].

Hairpin (PDB: 2OUE)

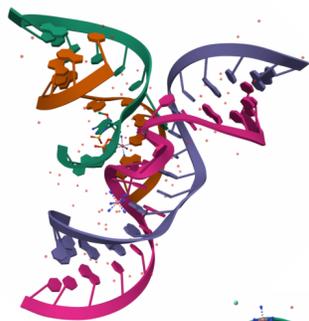

Twister (PDB: 4QJH)

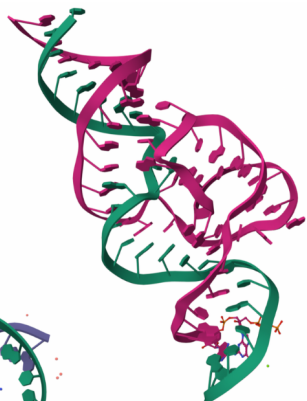

Hatchet (PDB: 6JQ5)

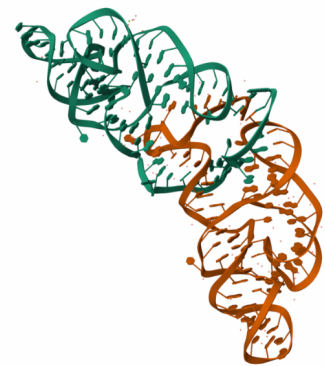

Pistol (PDB: 5KTJ)

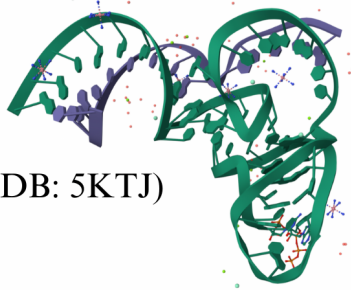

*Figure 1*. 3-D structure examples for four classes of ribozyme: hairpin, twister, pistol, and hatchet. Structures obtained from the RCSB PDB. No 3-D structure is available for hovlinc ribozymes, which were just discovered in 2021.

In recent years, deep learning has revolutionized the field of bioinformatics, especially in the study of proteins [7]. Alpha Fold, a deep learning model developed by DeepMind, has significantly enhanced our ability to predict protein structures from their amino acid sequences [8]. Other models focused on proteins include ESM-2 from Meta, ProtTrans, and I-TASSER

[9-11]. These breakthroughs have opened up new possibilities for predicting protein function from sequence, designing novel proteins with specific functions, and more important tasks. However, the application of deep learning to the study of RNA, including ribozymes, has been relatively unexplored.

Word2Vec is a landmark unsupervised learning algorithm developed for natural language processing (NLP) that learns to represent words as high-dimensional vectors, or embeddings, such that the semantic relationships between words are reflected in the relationships between their embeddings [12-13]. Word2Vec has been highly successful in NLP, enabling significant advances in tasks such as machine translation, sentiment analysis, and information retrieval [14-16]. One of the key advantages of Word2Vec is that it does not require labeled data, which can be difficult or expensive to curate in large quantities.

Applying Word2Vec to biological sequences is a promising direction for bioinformatics. Just as words in a sentence have a specific order and context that gives them meaning, the nucleotides in a ribozyme sequence have a specific order and context that influences their function. Like proteins, ribozyme structure is determined by sequence, suggesting that sequence information contains relevant information about ribozyme 3-D structure. By treating ribozyme sequences as "sentences" of nucleotides, or "words", Word2Vec can theoretically learn to represent these sequences as high-dimensional vectors. In this study, we split ribozyme sequences into 6-mers, which are the equivalent of words in regular Word2Vec. Different from Word2Vec, we use overlapping 6-mers instead of consecutive non-overlapping 6-mers.

The resulting ribozyme embeddings can then be used to study the relationships between different ribozymes, predict their functions, and classify them into different types. This natural language processing-inspired approach has the potential to significantly enhance our understanding of ribozymes and open up new avenues for research.

## Methods

Ribozyme sequences were retrieved from RNAcentral, a comprehensive database of noncoding RNA sequences [17]. Sequences with nucleotides other than A, T, C, and G were removed from consideration in order to control the vocabulary size of the Word2Vec model and eliminate non-standard sequence representations. Next, the bioinformatics tool CD-HIT was used to cluster sequences with greater than 75% sequence identity [18]. From each cluster, only one sequence was retained to represent the cluster. This was done to remove exact duplicates and near-duplicate sequences, which are common on the RNAcentral database. Near-duplicate sequences were removed to ensure sequence diversity and prevent any unwanted overfitting on overrepresented sequences. For instance, there could be several hundred near-identical ribozymes that would skew the learning of the Word2Vec model, especially on our relatively limited dataset. This was less of a problem for the default natural language-based Word2Vec which was trained on 100 billion words from Google News [19].

Each sequence was split into overlapping k-mers of six bases (6-mers). A k-mer is a substring of a sequence that is k nucleotides long. The k-mers were generated for each sequence in the dataset. These k-mers are essentially words and the sequence they come from is a sentence or document. We elected to use 6-mers because there are many possible 6-mers. Using 3-mers, as is common in bioinformatics, would only result in 64 possible words. Any given RNA sequence of enough length would contain many of them, by chance. It would potentially be difficult for the Word2Vec algorithm to learn to separate sequences with a vocabulary of only 64 words. For instance, Google's Word2Vec has a vocabulary of three million words and short phrases.

The vocabulary of the dataset was 4,096 (four possible nucleotides in each of six slots). The Word2Vec model was trained using the Gensim library in Python [20]. The model was trained on the 6-mers obtained from the sequences. The parameters for the Word2Vec model were set as follows: vector size was set to 128 or 256, window size was set to 5, minimum count was set to 1, and the number of worker threads was set to 4. The model was trained for 5 epochs. The window size is the number of 6-mers to the left and right that are considered to predict each 6-mer.

Since the Word2Vec model only provides embeddings for 6-mers, to calculate embeddings for a full sequence, the embedding for each constituent k-mer was calculated and the average was taken to get the sequence embedding. Averaging word embeddings is standard practice.

The model was tested on five different types of ribozymes: hatchet, pistol, hairpin, hovlinc, and twister sister. The sequences were retrieved in FASTA format from RNAcentral. The sequences for each type of ribozyme were loaded and preprocessed in the same way as the training data. However, CD-HIT filtering was not applied. The sequence embedding was calculated for each sequence in the test data.

| Ribozyme class | Number of samples |
|---|---|
| Hatchet | 257 |
| Pistol | 481 |
| Hairpin | 144 |
| Hovlinc | 88 |
| Twister Sister | 113 |
| Total | 1083 |

*Table 1.* Number of samples per ribozyme class in the dataset used to train and test the SVM.

The embeddings were then visualized using Principal Component Analysis (PCA). The PCA was performed on the embeddings and the first two principal components were plotted. The plot showed the separation of the five classes of ribozymes.

A Support Vector Machine (SVM) classifier was trained on the embeddings to classify the ribozymes. The data was split into training and test sets, with 50% of the data used for testing. The SVM classifier was trained on the training set and predictions were made on the test set. The performance of the classifier was evaluated using a classification report and a confusion matrix.

The quality of the clusters formed by the embeddings was evaluated using the Silhouette Score, Calinski-Harabasz Index, and Davies-Bouldin Index [21-23]. These metrics provide a quantitative measure of the quality of the clusters. Higher scores are better for the Silhouette Score and the Calinski-Harabasz Index. Meanwhile, lower scores are better for the Davies-Bouldin index.

The 128-D and 256-D models were saved and provided for batch inference on a web server. The web server was programmed in Python using the Streamlit framework.

# Results

There were 30,921 ribozyme sequences available in the RNAcentral database. Following CD-HIT clustering, 9,263 diverse sequences remained (sequence identity < 75%). The sequences were split into overlapping 6-mers, on which two Word2Vec models were trained. One model was trained to output 128-dimensional (128-D) embeddings and the other was trained to output 256-dimensional (256-D) embeddings. Principal component analysis showed that both embeddings were able to distinctly separate hatchet ribozymes, pistol ribozymes, and twister sister ribozymes.

Pistol ribozymes exhibited wide variance within the pistol cluster, suggesting that pistol ribozymes have relatively diverse structures and sequences. Meanwhile, hatchet ribozymes were tightly packed. For both the 128-D and 256-D embeddings, hovlinc and hairpin ribozymes were not strongly separated from each other, but were separated from the other three classes.

Overall, the two PCA plots are almost identical, except for some minute differences. Additionally, the Silhouette, Calinksi-Harabasz, and Davies-Bouldin scores were very similar, although the 256-D embeddings model slightly outperformed. This suggests that the 128-D embeddings capture most of the information that 256-D embeddings do.

128-dimensional embeddings PCA          256-dimensional embeddings PCA

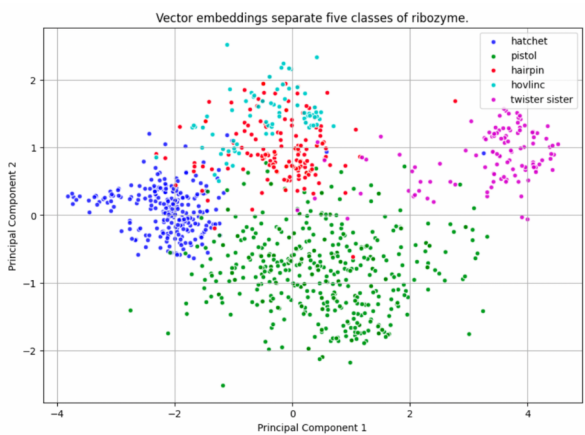 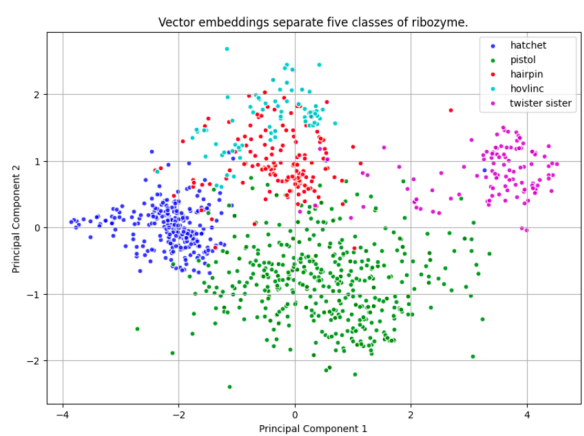

*Figure 2*. Side-by-side comparison of PCAs for the five ribozyme classes based on 128-D and 256-D embeddings. Each dot represents the projection of the embedding for a ribozyme. The PCA plots appear visually identical but upon closer inspection have slight differences. The five classes form clusters. However, the hairpin and hovlinc ribozymes overlap significantly. Legend: Purple: hatchet, Green: pistol, Red: hairpin, Blue: hovlinc, Pink: twister sister.

| Metric | 128-D embeddings | 256-D embeddings |
| --- | --- | --- |
| Silhouette Score | 0.331 | **0.339** |
| Calinski-Harabasz Index | 785.406 | **796.189** |
| Davies-Bouldin Index: | 1.449 | **0.851** |

*Table 2*. Comparison of common clustering metrics for the 128-D and 256-D embeddings.

Finally, multi-class Support Vector Machines were trained on embeddings for 1,083 ribozymes from the five classes. The SVMs trained on 128-D and 256-D embeddings performed similarly, both with an overall classification accuracy of 99% on the test set. Performance was robust across classes, despite moderate class imbalance. The 256-D model outperformed the 128-D model on classifying hovlinc and hairpin ribozymes. Interestingly, these are the same two classes of ribozyme that were not clearly separated by the PCA.

Apparently, the additional dimensionality of the 256-D embeddings potentially contained some information that was useful to the SVM in distinguishing hovlinc and hairpin ribozymes. However, the difference in accuracy is very small and the test set only contained around 500 ribozymes. A model classifying a single sample differently would influence the accuracy significantly. Finally, confusion matrices were generated to visualize the correct and incorrect predictions by class in a five-by-five grid. The confusion matrices showed that both SVMs got almost the same number of correct predictions in each class and tended to misclassify the same amount. For instance, the matrices reveal that both SVMs misclassified a hairpin ribozyme as a twister sister ribozyme.

| Classification Report for SVM Trained on 128-D embeddings | | | | |
|---|---|---|---|---|
| Class | Precision | Recall | F1-Score | Support |
| Hatchet | 0.98 | 0.98 | 0.98 | 129 |
| Pistol | 1 | 0.99 | 1 | 258 |
| Hairpin | 0.96 | 0.98 | 0.97 | 65 |
| Hovlinc | 1 | 0.98 | 0.99 | 43 |
| Twister Sister | 0.96 | 1 | 0.98 | 47 |
| Accuracy | | | 0.99 | 542 |
| Macro Avg | 0.98 | 0.99 | 0.98 | 542 |
| Weighted Avg | 0.99 | 0.99 | 0.99 | 542 |

*Table 3.* Classification report for the SVM trained on the 128-D embeddings

| Classification Report for SVM Trained on 256-D embeddings | | | | |
|---|---|---|---|---|
| Category | Precision | Recall | F1-Score | Support |
| Hatchet | 0.99 | 0.98 | 0.98 | 129 |
| Pistol | 1 | 1 | 1 | 258 |
| Hairpin | **0.97** | 0.98 | **0.98** | 65 |
| Hovlinc | 1 | **1** | **1** | 43 |
| Twister Sister | 0.96 | 1 | 0.98 | 47 |
| Accuracy | | | 0.99 | 542 |
| Macro avg | 0.98 | 0.99 | **0.99** | 542 |
| Weighted avg | 0.99 | 0.99 | 0.99 | 542 |

*Table 4*. Classification report for the SVM trained on the 256-D embeddings. Bolded scores indicate where the SVM outperformed the 128-D SVM.

Confusion Matrix for SVM trained on 128-D embeddings

Confusion Matrix for SVM trained on 256-D embeddings

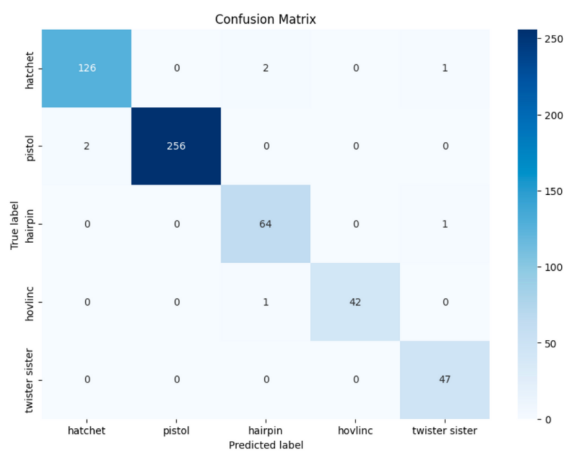
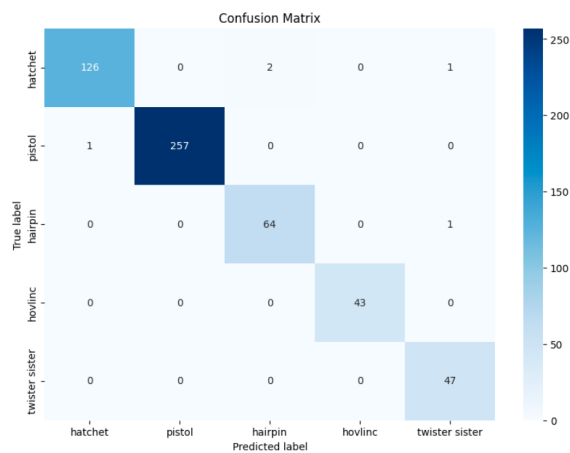

*Figure 3*. Confusion matrices for the predictions of the SVMs trained on the 128-D and 256-D embeddings.

# Conclusion

We demonstrate the application of Word2Vec, an unsupervised learning technique originally developed for natural language processing, to the field of bioinformatics, specifically for learning embeddings for ribozymes. Ribozymes are interesting because like proteins, they form distinct 3-D structures that are directly relevant to their function. Our model, Ribo2Vec, was trained on a diverse set of over 9,000 ribozymes, learning to map ribozyme sequences to 128 and 256-dimensional vector spaces. The resulting embeddings were found to contain meaningful information about the ribozymes, as evidenced by their ability to distinguish between five different classes of ribozymes. This was demonstrated through principal component analysis, which showed clear separation between different ribozyme classes. Furthermore, a simple SVM classifier trained on these embeddings showed promising results, achieving an overall classification accuracy of 99% on the test set.

Interestingly, the study found that the 256-dimensional embeddings behaved similarly to the 128-dimensional embeddings, suggesting that a lower dimension vector space is generally sufficient to capture the essential features of ribozymes. A limitation of the study was that some of the ribozyme sequences in the test set were included in training the Word2Vec model so there was some data contamination. Additionally, we only used around 9,000 ribozyme sequences to train the Word2Vec model which could be insufficient to learn a very good embedding. For comparison, Google's Word2Vec was trained on 100 billion words. Another limitation is that we only tested and implemented 6-mers. There could be a better length k-mer to use. Additionally, using variable length k-mers could potentially provide good results. Furthermore, adjusting the Word2Vec window size could change the quality of the embeddings. Also, there are many other classes of ribozymes such as hammerhead ribozymes and GlmS ribozymes that are potentially not well-represented by Ribo2Vec.

The results of this study open up avenues for RNA research, demonstrating the potential of Word2Vec for bioinformatics tasks like sequence classification and clustering. Future research could explore the use of Transformer-based methods to learn RNA embeddings, which could capture long-range interactions between nucleotides. Additionally, larger, more diverse datasets of RNA sequences could be used. The approach could be applied to learn embeddings for other types of noncoding RNAs, such as miRNAs and snoRNAs. This could further enhance our understanding of noncoding RNAs and their functions. A similar approach could be used to learn embeddings for epitopes for computational vaccine design purposes (24-25). In conclusion, this study has shown that deep learning techniques, such as Word2Vec, can be effectively applied to RNA sequences, providing a novel approach to understanding these important biological molecules.

## Data Availability

The Ribo2Vec models are provided on a web server: https://ribo2vec.sites.stanford.edu/
The data used to train the models was retrieved from RNAcentral, a public database of noncoding RNAs.